\documentclass{article}

\usepackage{arxiv}

\usepackage[utf8]{inputenc} 
\usepackage[T1]{fontenc}    
\usepackage{nicefrac}       
\usepackage{microtype}      
\usepackage{lipsum}		
\usepackage{natbib}
\usepackage{doi}

\usepackage{times}
\usepackage{epsfig}
\usepackage{graphicx}
\usepackage{amsfonts}       
\usepackage{amsmath}
\usepackage{amssymb}
\usepackage{amsthm}
\usepackage{mathtools}
\usepackage{multirow}
\usepackage{booktabs} 
\usepackage{subfigure}
\usepackage{xcolor,colortbl}
\usepackage{adjustbox}
\usepackage{url}

\usepackage{tikz}
\def\checkmark{\tikz\fill[scale=0.4](0,.35) -- (.25,0) -- (1,.7) -- (.25,.15) -- cycle;} 

\definecolor{Gray}{gray}{0.85}
\definecolor{LightCyan}{rgb}{0.88,1,1}

\title{Enhancing Low-resolution Face Recognition with Feature Similarity Knowledge Distillation}

\author{Sungho Shin \\
	School of Integrated Technology (SIT)\\
	Gwangju Institute of Science and Technology (GIST)\\
	\texttt{hogili89@gm.gist.ac.kr} \\
	\And
	Yeonguk Yu \\
	School of Integrated Technology (SIT)\\
	Gwangju Institute of Science and Technology (GIST)\\
	\texttt{yeon\_guk@gm.gist.ac.kr} \\
	\And
	Kyoobin Lee \\
	School of Integrated Technology (SIT)\\
	Gwangju Institute of Science and Technology (GIST)\\
	\texttt{kyoobinlee@gist.ac.kr}
}

\renewcommand{\shorttitle}{Feature Similarity Knowledge Distillation}

\begin{document}
\maketitle

\begin{abstract}

In this study, we introduce a feature knowledge distillation framework to improve low-resolution (LR) face recognition performance using knowledge obtained from high-resolution (HR) images. The proposed framework transfers informative features from an HR-trained network to an LR-trained network by reducing the distance between them. A cosine similarity measure was employed as a distance metric to effectively align the HR and LR features. This approach differs from conventional knowledge distillation frameworks, which use the $L_p$ distance metrics and offer the advantage of converging well when reducing the distance between features of different resolutions. Our framework achieved a 3\% improvement over the previous state-of-the-art method on the AgeDB-30 benchmark~\cite{agedb} without bells and whistles, while maintaining a strong performance on HR images. The effectiveness of cosine similarity as a distance metric was validated through statistical analysis, making our approach a promising solution for real-world applications in which LR images are frequently encountered. The code and pretrained models are publicly available on \url{https://github.com/gist-ailab/feature-similarity-KD}.

\end{abstract}


\section{Introduction}
\begin{figure}
\centering
\includegraphics[height=7cm]{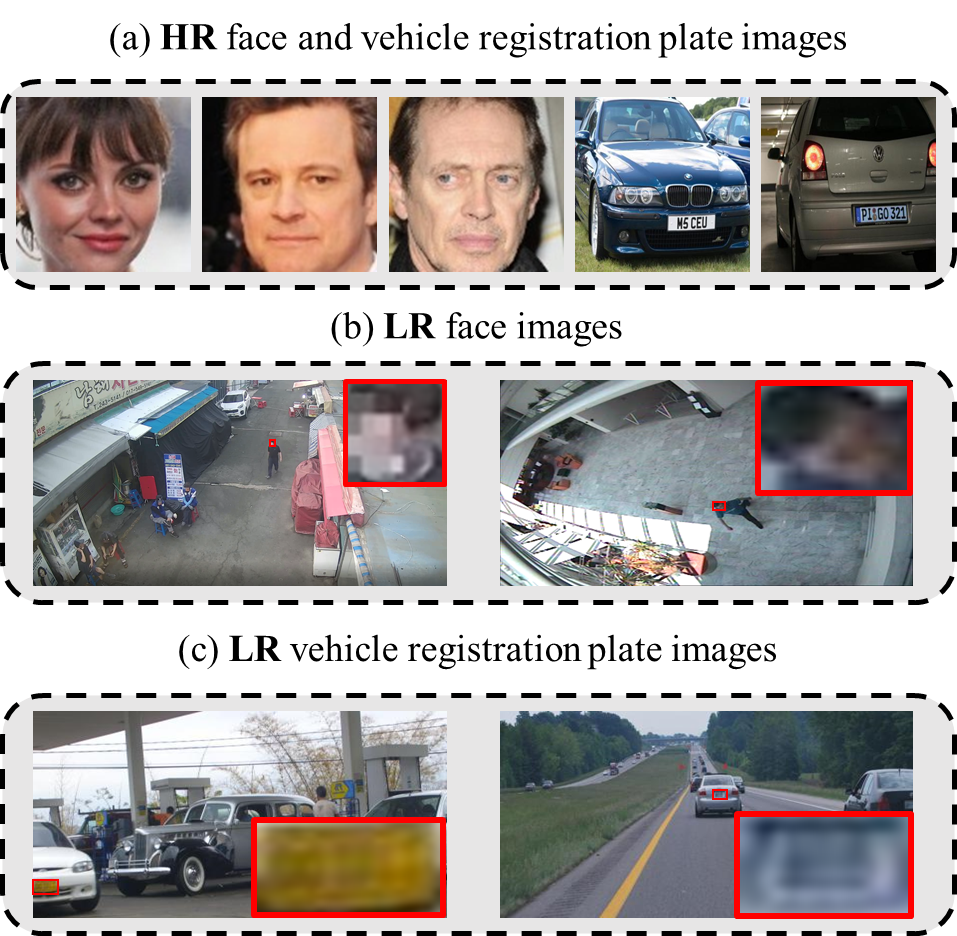}
\caption{Example images of high-resolution (HR) and low-resolution (LR) samples of faces and vehicle registration plates. While various approaches have been proposed for face identity and digit recognition in HR images, as shown in (a), many face and registration plate images are frequently captured in low-resolution circumstances, as depicted in (b) and (c).}
\label{fig:intro}
\end{figure}

Deep learning has been widely utilized for face recognition tasks because of its superior performance~\cite{arcface,fr_2,fr_3,fr_1}. A recent study reported accuracy rates of 99\% on the MegaFace benchmark, which involves identifying a given image among one million distractors~\cite{megaface}. However, in real-world situations, face images may be captured under a variety of challenging conditions, such as low-resolution (LR), low-illumination, and small size, which can significantly degrade the performance of deep-learning models. These conditions are not well represented in existing face recognition benchmarks, such as CASIA~\cite{casia} and MegaFace~\cite{megaface}, which comprise high-resolution (HR) images with a resolution of at least 112 pixels in width and height. LR images, on the other hand, may contain as few as 64 pixels (8 $\times$ 8) in extreme cases as depicted in Figure \ref{fig:intro}.

The resolution of face images can affect the accuracy of face recognition models. When the resolution is degraded, the models may not capture sufficient detail for accurate recognition, owing to the loss of spatial information. When the high-accuracy margin network (ArcFace~\cite{arcface}) was tested on LR datasets (\textit{e.g.,} IJB-B~\cite{ijbb}, IJB-C~\cite{ijbc}, and TinyFace~\cite{tinyface}), a significant degradation was observed (accuracies below 70\%). Kim \textit{et al.} noted that significant resolution degradation can make images unidentifiable, causing models to rely on other features, such as hairstyle and clothing, which can bias model training~\cite{adaface}. Recent studies have addressed this by imposing different weights based on the image quality~\cite{lq_1,adaface,lq_2,lq_3}. For example, AdaFace demonstrated that the feature norm is positively correlated with the image quality and designed an adaptive margin based on the feature norm to emphasize high-quality hard samples and de-emphasize low-quality ones~\cite{adaface}. These approaches aim to select informative images from a large number of mixed-resolution face images, but do not focus on extracting informative features from LR images.

 Other studies~\cite{qualnet,Massoli2020,askd} employed the knowledge distillation (KD) framework to extract informative features from LR images based on guidance from a teacher network trained on HR images. These approaches encourage the intermediate features produced by the face recognition network at lower resolutions to resemble those produced by a teacher network trained using HR images. For example, QualNet-LM~\cite{qualnet} uses a face recognition network with a decoder structure trained on HR images as the teacher network. To extract informative features from LR images, they added a reconstruction loss to regulate the student network's LR features such that they could be decoded into HR images when passed through the decoder of the teacher network. A-SKD~\cite{askd} uses KD to transfer well-constructed attention maps from the HR network to the attention maps of the LR network, achieving state-of-the-art (SOTA) results on various LR face recognition benchmarks. However, these methods are limited to specific frameworks (\textit{e.g.,} decoder, and attention modules) to align the features of the teacher and student networks at different resolutions.

 In this study, we present a simple approach for knowledge transfer between an HR and LR face recognition network using feature similarity knowledge distillation (F-SKD). We aim to align the features produced by both networks and transfer the knowledge obtained from the HR network to the LR network without additional modules. Through experiments and statistical analyses, we found that cosine similarity is key to reducing the distance between features of different resolutions, outperforming traditional KD frameworks that use $L_p$ distance metrics. Our approach outperformed previous SOTA results on LR face recognition tasks by 3\% for AgeDB-30~\cite{agedb} and 1.55\% for TinyFace~\cite{tinyface} without bells and whistles. Even in the LR digit classification task, our approach outperformed the previous methods by 1.73\% on the SVHN benchmark~\cite{svhn}. F-SKD is a promising solution for LR face recognition applications in the real-world, owing to its efficiency and ease of implementation.

 \section{Related Works}
\textbf{Face Recognition.} Face recognition networks must differentiate between a large number of face images belonging to different identities in open-set environments~\cite{arcface, sphereface, cosface}. The traditional softmax loss function lacks a margin between the decision boundaries of different classes, which can result in incorrect classifications owing to the small perturbations. To address this issue, several margin-based face recognition losses have been proposed to increase the similarity of intra-class samples and dissimilarity of inter-class samples~\cite{arcface, sphereface, cosface}. SphereFace~\cite{sphereface} introduced the angular-softmax (A-softmax) loss, which multiplies the margin ($m$) with the angle ($\theta$) between the feature vector ($\mathbf{x}$) and weight vector ($\mathbf{W}_{y}$) of the target class ($y$). However, the margin of A-softmax loss is angle-dependent and disappears when the angle becomes zero. CosFace~\cite{cosface} avoids this problem by subtracting $m$ from $cos(\theta)$. ArcFace~\cite{arcface} proposed adding $m$ directly to $\theta$ resulting in $cos(\theta + m)$, which provides a constant linear angular margin throughout the interval, whereas CosFace has only a nonlinear angular margin. These angular margin-based loss functions have become the standard for face recognition owing to their superior performance and simplicity of implementation.

\textbf{Low-resolution Face Recognition.}
LR face recognition studies have focused on leveraging the knowledge obtained from HR images to improve LR face recognition performance using KD frameworks~\cite{qualnet,Massoli2020,askd}. A teacher network trained on HR images can extract more informative features for recognition, which can then be transferred to a student network trained on LR images. Massoli \textit{et al.} demonstrated that reducing the feature distance between HR and LR networks significantly enhances LR recognition performance~\cite{Massoli2020}. Kim \textit{et al.} proposed the QualNet-LM framework that employs a face recognition network with a decoder structure trained on HR images as the teacher network~\cite{qualnet}. To effectively extract information from the LR images, a reconstruction loss was incorporated to regulate the LR features of the student network, enabling them to be decoded into HR images when passed through the decoder of the teacher network. Shin \textit{et al.} suggested utilizing KD to transfer well-designed attention maps from an HR network to the attention maps of an LR network, obtaining SOTA results on various LR face recognition benchmarks~\cite{askd}. However, existing studies are limited to specific frameworks, such as decoders and attention modules, to align features of varying resolutions. This study investigates an efficient KD approach that can align different resolution features without being restricted to any specific frameworks, achieving superior performance.

\section{Proposed Approach}\label{methods}
This section first introduces the CosFace angular margin face recognition network~\cite{cosface}, which was utilized in the previous KD method QualNet-LM (Sec.\ref{sec:method1}). We then describe the F-SKD approach, which increases the similarity between the teacher and student network features (Sec.\ref{sec:method2}). Finally, we present the \textit{t}-test~\cite{ttest} and Pearson's correlation analysis~\cite{correlation} to evaluate the effectiveness of the proposed cosine similarity-based distance metric (Sec.\ref{sec:method3}).

\subsection{Recap: CosFace}
\label{sec:method1}
For the recognition task, softmax loss is widely utilized to separate features from different classes by maximizing the posterior probability of the labels. Given an input feature vector of $i$th sample $\mathbf{x}_i \in \mathbb{R}^d$ with its corresponding label $y_i$, the softmax loss can be formulated as

\begin{equation} 
\label{eq:softmax}
L_{softmax} = -\frac{1}{N} \sum_{i=1}^{N}log\frac{e^{\mathbf{W}^T_{y_i}\mathbf{x}_i}}{\sum_{j=1}^{n}e^{\mathbf{W}^T_j\mathbf{x}_i}} , 
\end{equation}

$N$ and $n$ are the batch size and number of classes, respectively, $\mathbf{W}_j \in \mathbb{R}^d$ denotes the $j$th column of the last fully connected layer's weight $\mathbf{W} \in \mathbb{R}^{d \times n}$. For simplicity, we set the bias terms to zero.

 For effective feature learning, previous studies~\cite{sphereface,cosface} have fixed $\lvert\lvert \mathbf{W}_j \rvert\rvert_2 = 1$ and $\lvert\lvert \mathbf{x}_i \rvert\rvert_2 = s$ to ensure that the posterior probability relies only on the cosine angle between the feature and fully connected layer's weights. We can then rewrite the term $\mathbf{W}^T_j\mathbf{x}_i = \lvert\lvert \mathbf{W}_j \rvert\rvert_2 \lvert\lvert \mathbf{x}_i \rvert\rvert_2 cos(\theta_j) = scos(\theta_j)$, where $\theta$ denotes the angle between $\mathbf{x}_i$ and $\mathbf{W}_j$. Finally, CosFace~\cite{cosface} introduced the cosine margin ($m$) to strengthen the discrimination between the cosine angles of different classes: $cos(\theta_i) - m > cos(\theta_j)$ and $cos(\theta_j) - m > cos(\theta_i)$, where $i$ and $j$ denote different classes. CosFace~\cite{cosface} loss can be formulated as:

\begin{equation} 
\label{eq:cosface}
L_{cosface} = -\frac{1}{N} \sum_{i=1}^{N}log\frac{e^{s(cos(\theta_{y_i}) - m)}}{e^{s(cos(\theta_{y_i}) - m)} + \sum_{j \ne y_i}e^{scos(\theta_j)}} . 
\end{equation}

\subsection{Feature Similarity Knowledge Distillation}
\label{sec:method2}
\begin{figure}
\centering
\includegraphics[height=5cm]{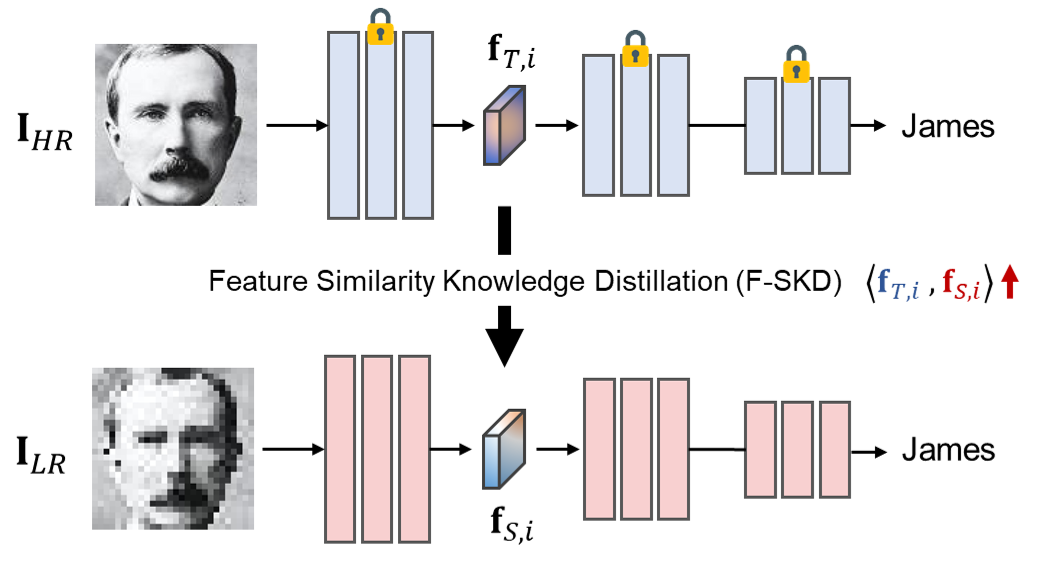}
\caption{Proposed F-SKD framework where the HR network guides the LR network to extract informative features, such as eyes, nose, and lips by improving the similarity between their respective features. The figure shows F-SKD only for the first block.}
\label{fig:method}
\end{figure}

To extract informative features from LR images, we utilized a KD framework that uses paired HR-LR samples as input (Figure \ref{fig:method}). This framework aims to reduce the feature difference between the corresponding locations in the paired samples, thereby enabling the LR network to extract richer features at a level comparable to that of the HR network. The $L_p$ distance function is employed for the KD loss in general object classification tasks~\cite{kd_1, rkd, fitnet, kd_2}. However, in this study, we proposed the use of cosine similarity-based distance function. 

 Our loss function was designed to minimize the angle between the feature vectors of the teacher and student networks. Previous studies have shown that feature norm is positively correlated with image quality, and an adaptive margin function has been introduced to emphasize samples with high-quality features based on their norm. Following this, we transfer only the direction component of the feature vector (the normalized feature vector) from the teacher network, instead of trying to match the norm component of the student network's features to those of the teacher's. This approach enables the LR network to learn where to focus among the feature elements for extracting richer information guided by the HR network while avoiding bias towards other characteristics such as clothing and hairstyle.
 
 Let $\mathbf{f}_i = \mathcal{H}_i(\mathbf{I})$ be the intermediate feature outputs from the \textit{i}th block of the CNN for the given image $\mathbf{I}$. The cosine similarity-based KD loss can be described as follows:

\begin{equation}
\label{eq:distill_loss}
\begin{split}
L_{distill} &= \frac{1}{L} \sum_{i=1}^{L}(1 - \langle \mathbf{f}_{T,i}, \mathbf{f}_{S,i} \rangle) \\
       &= \frac{1}{L} \sum_{i=1}^{L}(1 - \frac{\mathbf{f}_{T,i}}{\lVert \mathbf{f}_{T,i} \rVert_2} \cdot
      \frac{\mathbf{f}_{S,i}}{\lVert \mathbf{f}_{S,i} \rVert_2}) , 
\end{split}
\end{equation}

where $\langle \cdot, \cdot \rangle$ indicates the cosine similarity operation, $L$ denotes the number of blocks employed for the distillation, and $\mathbf{f}_{T,i}$ and $\mathbf{f}_{S,i}$ denote the \textit{i}th CNN layer's features from the teacher and student networks, respectively.

The total loss for the LR face recognition network is the sum of the face recognition loss (\ref{eq:cosface}), CosFace is utilized in this study, and the cosine similarity-based KD loss (\ref{eq:distill_loss}) weighted by the factor ($\lambda_{distill}$). 

\begin{equation} 
\label{eq:total_loss}
L_{total} = L_{cosface} + L_{distill} * \lambda_{distill}  .
\end{equation}

\subsection{Statistical Analysis}
\label{sec:method3}
In this study, we propose using similarity loss (\ref{eq:distill_loss}) to distill only the directional components of the teacher network's features in the student network. To assess the effectiveness of this approach, we used a \textit{t}-test to determine if the norm of the student network's features is statistically the same as that of the teacher network after distillation and performed Pearson's correlation analysis to determine if the direction of the student network's features aligns with that of the teacher network after distillation.

\textbf{\textit{t}-test.} \textit{t}-test~\cite{ttest} is a statistical hypothesis test that tests the null hypothesis that the mean difference between two groups of samples is equal to zero, against the alternative hypothesis that the mean difference is not zero. This is based on the assumption that data are normally distributed. To validate the null hypothesis, $\mu_x - \mu_y = 0$, where $\mu$ is the mean statistic of each group, the \textit{t}-test uses the \textit{t}-statistic, which can be calculated as follows:

\begin{equation}
\label{eq:ttest}
t = \frac{\mu_x - \mu_y}{\sqrt{(s_x^2 + s_y^2) / n}} ,
\end{equation}

where $s_x$ and $s_y$ are the standard deviations of the two groups and $n$ is the number of observations in each group.

The calculated \textit{t}-value is then compared to the critical value ($c$), which is determined based on the significance level ($\alpha$). If the calculated \textit{t}-value is greater than the critical value, then we reject the null hypothesis. The critical value $c$ can be determined by the condition $\int_{c}^{\infty} f(t)dt = \alpha$, where $f(t)$ represents the probability density function of the Student's \textit{t}-distribution. In the study, $\alpha = 0.01$.

\textbf{Pearson's Correlation Analysis.} The Pearson's correlation analysis~\cite{correlation} measures the linear relationship between two continuous variables and their strengths. The correlation coefficient was calculated using the following equation:

\begin{equation}
\label{eq:pearson_correlation}
r = \frac{\sum_{i=1}^n (x_i - \mu_x)(y_i - \mu_y)}{\sqrt{\sum_{i=1}^n (x_i - \mu_x)^2}\sqrt{\sum_{i=1}^n (y_i - \mu_y)^2}} ,
\end{equation}

where $x_i$ and $y_i$ are the values of the two variables for the $i$th observation, $\mu_x$ and $\mu_y$ are the means of the two variables, and $n$ is the number of observations.

The range of the coefficient is between -1 and 1. If the coefficient is 1, it means there is a perfect positive linear relationship between the variables. Conversely, if the coefficient is -1, it indicates a perfect negative linear relationship between the variables. On the other hand, if the coefficient is 0, it suggests that there is no linear relationship between the variables.

\section{Experiments}\label{experiments}
\subsection{Task}
Face recognition is the process of identifying individuals in facial images. There are two main types of face recognition methods: face verification and face identification. In face verification, the system compares two facial images to confirm whether they belong to the same person using a 1:1 comparison. In face identification, the system attempts to identify the person in a probe image by comparing it to a larger set of images, known as the gallery set, through a 1:N comparison. Digit classification, in contrast, is the task of recognizing numbers in images containing a single digit. The system classifies these images as 0--9.

\subsection{Datasets}
\textbf{Face Recognition.} We used the CASIA~\cite{casia} dataset, which contains 0.5 million face images and 10K identities, to train our face recognition model. To create HR-LR paired face images, we downsampled the resolution of the images from CASIA by reducing their size and then enlarged them to their original size using bilinear interpolation. For the evaluation, we used the AgeDB-30~\cite{agedb} and TinyFace~\cite{tinyface} datasets. AgeDB-30 had 16,516 images with 570 subjects for the 1:1 verification task, and TinyFace had 169,403 LR face images with 5,139 subjects for the 1:N identification task. TinyFace is a well-known LR face recognition benchmark captured in real-world settings.

\textbf{Digit Classification.} We used the SVHN~\cite{svhn} dataset, which has character-level digit images (0 -- 9) obtained from house numbers in Google Street View images. It contained 73,257 and 26,032 digit images for training and evaluation, respectively.

\subsection{Settings}
 \textbf{Evaluation Protocol.} We downsampled images from CASIA with ratios of 2$\times$, 4$\times$, and 8$\times$ in accordance with recent LR face recognition studies~\cite{qualnet,askd}. Here, 1$\times$ denotes the HR images. Two benchmark settings were employed for the evaluation: (1) a single-resolution, where a student network is trained to recognize images from one resolution, and (2) multiple-resolutions, where a student network is trained to recognize images from multiple-resolutions. Because AgeDB-30 has a similar resolution to CASIA, networks trained on downsampled CASIA images were validated on AgeDB-30 downsampled images with matching ratios and validated in both single- and multiple-resolution settings. In contrast, TinyFace, a real-world LR benchmark, comprises face images with different resolutions; therefore, networks trained on multiple-resolution settings were validated using TinyFace. For digit classification, we compared the classification accuracy on 4$\times$ downsampled SVHN images by following a single-resolution evaluation setting.

\textbf{Backbone.} Our method employs the KD framework, which uses a teacher-student network. To ensure fair comparison with previous studies, we used the same backbone structures as those used in previous studies. We utilized the iResNet50~\cite{arcface} backbone for both the teacher and student networks compared with FitNet~\cite{fitnet} and RKD~\cite{rkd}. When compared with QualNet-LM~\cite{qualnet}, we used iResNet50 for the student network and iResNet50 combined with a decoder module (iRevNet-300~\cite{irevnet}) for the teacher network because QualNet-LM requires a decoder module for the teacher network. Lastly, when compared with A-SKD~\cite{askd}, we used iResNet50 with CBAM modules for both the teacher and student networks, as A-SKD requires attention modules to extract and transfer the attention maps. For digit classification, we replaced iResNet50 with a ResNet50~\cite{resnet50} backbone, which is commonly used in classification tasks.

\textbf{Implementation Details.} 
In this study, we followed the common practices proposed by \cite{arcface} for preprocessing the face recognition dataset. This process included using the MTCNN~\cite{mtcnn} face detector to locate and crop the face region and then resizing the cropped image to 112 $\times$ 112 pixels using bilinear interpolation. To ensure a fair comparison, all methods were re-implemented using the CosFace~\cite{cosface} margin and iResNet50 backbone by following the settings of the previous SOTA method (QualNet-LM~\cite{qualnet}). The learning rate was set to 0.1 initially and decreased by a factor of 10 at 18K, 28K, 36K, and 44K iterations. The SGD optimizer was used with a batch size of 256. The training process finished after 47K iterations. For digit classification, the learning rate was set to 0.01 initially and decreased by a factor of 10 at 30, 60, and 80 epochs. The SGD optimizer was used with a batch size of 64. The training process was completed after 90 epochs. For the hyper-parameter search ($\lambda_{distill}$), the CFP-FP~\cite{cfp} face dataset was used as an external validation set and a random search was conducted. Through the search, $\lambda_{distill} = 5$ for the face recognition task. We also employed the same weight factor for the digit classification task and it generally achieved superior performance.

\section{Results and Discussion}
\subsection{Low-resolution Face Recognition}
\textbf{Evaluation on AgeDB-30.} After training a face recognition network using downsampled images from the CASIA dataset, we evaluated the performance of the network using downsampled images from the AgeDB-30 dataset. First, we compared our approach to previous studies by following a single-resolution setting (Table \ref{result_agedb_single}). The F-SKD outperformed the previous SOTA methods in all cases (2$\times$, 4$\times$, and 8$\times$). For the most extreme case (8$\times$), F-SKD achieved 3.0\% and 0.8\% improvements compared with QualNet-LM~\cite{qualnet} and A-SKD~\cite{askd}, respectively. Second, when evaluated using multiple-resolution settings, F-SKD outperformed the other methods on average (Table \ref{result_agedb_multi}). Although A-SKD slightly outperformed F-SKD at 8$\times$ downsampled images (+0.001\%), it showed significant degradation in HR recognition performance compared to the teacher network (-2.1\%). In contrast, our approach maintained superior HR recognition performance while significantly improving LR recognition performance.


\begin{table}[t]
\caption{Single-resolution evaluation results of proposed distillation approaches on AgeDB-30. (a) Verification accuracy of FitNet, RKD, QualNet-LM, and F-SKD using iResNet50 as the student network. (b) Verification accuracy of A-SKD and F-SKD using iResNet50+CBAM as the student network. Base refers to a network that has not undergone any KD methods. The term Dec denotes the decoder structure utilized in the QualNet-LM. The highest scores among the comparisons are shown in \textbf{bold}.}
\vskip 0.1in
\label{result_agedb_single}
\centering
\resizebox{0.6\textwidth}{!}{%
\begin{tabular}{cccc}
\toprule
\textbf{Resolution} & \textbf{Type} & \textbf{Teacher} & \textbf{ACC (\%)} \\ \midrule

\rowcolor{Gray}
(a) &&& \\ \midrule

\multirow{2}{*}{112 $\times$ 112} & \multirow{2}{*}{Base} & iResNet50 & 92.63 \\ \cmidrule(l){3-4} 
    &  & iResNet50+Dec & 93.03 \\ \midrule
\multirow{6}{*}{56 $\times$ 56} & Base & - & 91.35 \\ \cmidrule(l){2-4} 
& FitNet~\cite{fitnet} & \multirow{3}{*}{iResNet50} & 91.57  \\
& RKD~\cite{rkd} &  &  92.05 \\
& F-SKD (Ours) & & \textbf{92.52} \\ \cmidrule(l){2-4}
& QualNet-LM~\cite{qualnet} & \multirow{2}{*}{iResNet50+Dec} & 91.37 \\
& F-SKD (Ours) & & \textbf{92.10}  \\ \midrule
\multirow{6}{*}{28 $\times$ 28} & Base & - & 85.68 \\ \cmidrule(l){2-4} 
& FitNet~\cite{fitnet} & \multirow{3}{*}{iResNet50} & 86.05  \\
& RKD~\cite{rkd} &  &  86.02 \\
& F-SKD (Ours) & & \textbf{87.42} \\ \cmidrule(l){2-4}
& QualNet-LM~\cite{qualnet} & \multirow{2}{*}{iResNet50+Dec} & 86.45 \\
& F-SKD (Ours) & & \textbf{87.68}  \\ \midrule
\multirow{6}{*}{14 $\times$ 14} & Base & - & 73.67 \\ \cmidrule(l){2-4} 
& FitNet~\cite{fitnet} & \multirow{3}{*}{iResNet50} & 74.73  \\
& RKD~\cite{rkd} &  &  74.30 \\
& F-SKD (Ours) & & \textbf{75.97} \\ \cmidrule(l){2-4}
& QualNet-LM~\cite{qualnet} & \multirow{2}{*}{iResNet50+Dec} & 74.33 \\
& F-SKD (Ours) & & \textbf{76.55}  \\ \midrule

\rowcolor{Gray}
(b) &&& \\ \midrule

112 $\times$ 112 & Base & iResNet50+CBAM & 93.12 \\ \midrule
\multirow{3}{*}{56 $\times$ 56} & Base & - & 90.78 \\ \cmidrule(l){2-4}
  & A-SKD~\cite{askd} &  \multirow{2}{*}{iResNet50+CBAM} & 91.62 \\
  & F-SKD (Ours) &  & \textbf{91.70} \\ \midrule
\multirow{3}{*}{28 $\times$ 28} & Base & - & 85.90 \\ \cmidrule(l){2-4}
  & A-SKD~\cite{askd} &  \multirow{2}{*}{iResNet50+CBAM} & 86.35 \\
  & F-SKD (Ours) &  & \textbf{87.53} \\ \midrule
\multirow{3}{*}{14 $\times$ 14} & Base & - & 73.62 \\ \cmidrule(l){2-4}
  & A-SKD~\cite{askd} &  \multirow{2}{*}{iResNet50+CBAM} & 74.78 \\
  & F-SKD (Ours) &  & \textbf{75.35} \\ \bottomrule

\end{tabular}
}
\end{table}

\begin{table*}[t]
\caption{Multiple-resolution evaluation results of proposed distillation approaches on AgeDB-30. (a) Verification accuracy of QualNet-LM and F-SKD on four different resolutions (112$\times$112, 56$\times$56, 28$\times$28, and 14$\times$14) using iResNet50 as the student network. (b) Verification results of A-SKD and F-SKD using iResNet50+CBAM as the student network.}
\vskip 0.1in
\centering
\label{result_agedb_multi}
\resizebox{\textwidth}{!}{%
\begin{tabular}{cccccccc}
\toprule

\textbf{Resolution} & \textbf{Type} & \textbf{Teacher} & \textbf{112 $\times$ 112 (\%)} & \textbf{64 $\times$ 64 (\%)} & \textbf{28 $\times$ 28 (\%)} & \textbf{14 $\times$ 14 (\%)} & \textbf{Average (\%)} \\ \midrule

\rowcolor{Gray}
(a) & &  & & & & & \\ \midrule
 
112 $\times$ 112 & Base & iResNet50+Dec & 93.03 & - & - & - & -\\ \midrule
\multirow{3}{*}{All} & Base & - & 90.27 & 89.63 & 85.58 & 74.58 & 85.02 \\ \cmidrule(l){2-8}
 & QualNet-LM~\cite{qualnet} & \multirow{2}{*}{iResNet50+Dec} & 92.77 & 91.85 & \textbf{87.35} & 75.57 & 86.89 \\ 
 & F-SKD (Ours) &  & \textbf{92.90} & \textbf{91.88} & \textbf{87.35} & \textbf{76.02} & \textbf{87.04} \\ \midrule

\rowcolor{Gray}
 (b) & &  & & & & & \\ \midrule

112 $\times$ 112 & Base & iResNet50+CBAM & 93.12 & - & - & - & - \\ \midrule
\multirow{4}{*}{All} & Base & - & 90.57 & 89.95 & 85.60 & 74.32 &  85.11 \\ \cmidrule(l){2-8}
 & A-SKD~\cite{askd} & \multirow{2}{*}{iResNet50+CBAM} & 91.12 & 90.35 & 86.03 & \textbf{75.32} & 85.71 \\ 
 & F-SKD (Ours) &  & \textbf{92.58} & \textbf{91.50} & \textbf{86.57} & 75.23 & \textbf{86.47} \\ \bottomrule
 
\end{tabular}
}
\end{table*}

\textbf{Evaluation on TinyFace.} TinyFace~\cite{tinyface} is the most widely used LR benchmark obtained from real-world scenarios. Because the images from TinyFace have diverse resolutions, we validated our approach using multiple-resolution settings (Table \ref{result_tinyface}). Similar to the AgeDB-30 results, F-SKD achieved 1.55\% and 3.10\% increased identification accuracy compared with QualNet-LM and A-SKD, respectively. From the results, we demonstrated that increasing the similarity between the HR and LR network features through the teacher-student framework is a simple but effective solution for efficiently incorporating multi-scale features into a single network.

\begin{figure*}
\centering
\includegraphics[height=7.2cm]{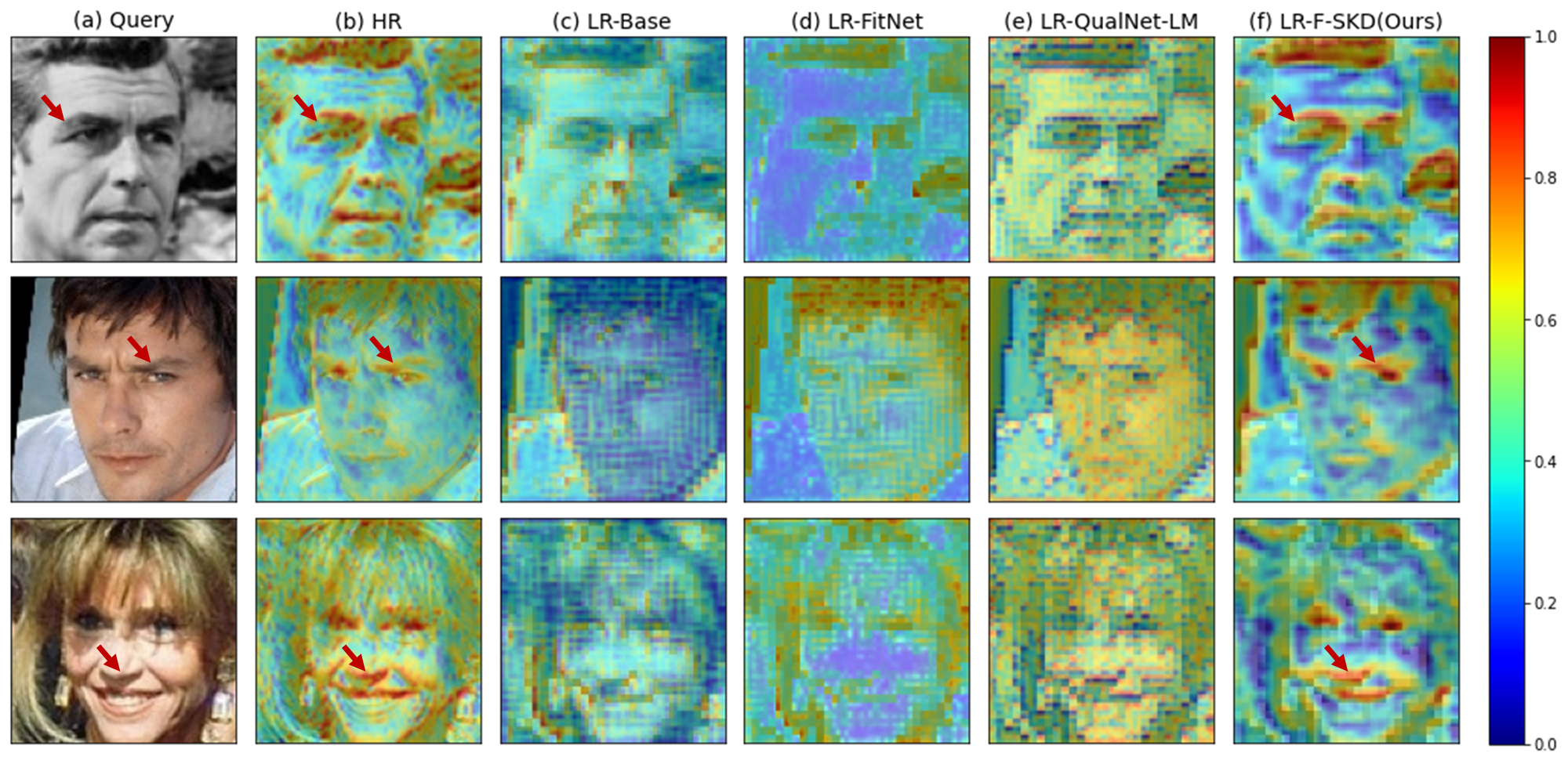}
\caption{Visualization of self-attention maps from the first block of HR and LR networks using different distillation approaches. After applying our approach, LR network focused on fine-grained features such as eyebrows and lips, similar to the HR network, as indicated by the red arrow. The colors red and blue represent high and low attention, respectively. Facial images used in the visualization are sourced from the AgeDB-30.}
\label{fig:attention}
\end{figure*}

\begin{table}[t]
\caption{Results of proposed distillation approaches on the real-world LR face identification benchmark, TinyFace.}
\vskip 0.1in
\centering
\label{result_tinyface}
\resizebox{0.65\textwidth}{!}{%
\begin{tabular}{cccc}
\toprule
\textbf{Type} & \textbf{Student} & \textbf{Teacher} & \textbf{ACC (\%)} \\ \midrule
Base & \multirow{3}{*}{iResNet50} & - & 57.30 \\ \cmidrule(l){1-1} \cmidrule(l){3-4}
QualNet-LM~\cite{qualnet} &   & \multirow{2}{*}{iResNet50+Dec} & 58.56 \\
F-SKD (Ours) & &  & \textbf{59.47} \\ \midrule
Base & \multirow{3}{*}{iResNet50+CBAM} & - & 54.94 \\ \cmidrule(l){1-1} \cmidrule(l){3-4}
A-SKD~\cite{askd} &   & \multirow{2}{*}{iResNet50+CBAM} & 55.42 \\
F-SKD (Ours) & &  & \textbf{57.14} \\ \bottomrule
\end{tabular}
}
\vskip -0.1in
\end{table}

\textbf{Visualization of Self-attention Map.} 
The features of the first block are visualized in Figure \ref{fig:attention} through a self-attention map~\cite{at}. To generate this map, channel-wise average pooling was applied to the features and resulting values were normalized to a range between 0 and 1. This visualization highlights the regions that are most relevant for face recognition in both the HR and LR networks. An HR network focuses on facial features, such as the eyes, nose, and lips, which are crucial for successful recognition~\cite{s2ld}. However, the LR network fails to capture these details owing to the loss of spatial information, leading to a degraded performance.

Unlike previous distillation methods, such as FitNet~\cite{fitnet} and QualNet-LM~\cite{qualnet}, our approach enhances the LR network's self-attention maps to match the HR network's attention patterns. Remarkably, our method extracts fine-grained details, such as wrinkle-like structures from limited visual information. This ability to focus on highly detailed facial features is a significant improvement over previous methods and a key factor in achieving SOTA performance.

\subsection{Effectiveness of Similarity Distillation}
To incorporate multi-scale features into a single network, we need to define the target object for KD and its distance measures. FitNet~\cite{fitnet} distills all features from the teacher network into student features via $L_2$ distance measures. However, it did not exhibit a significant improvement in LR recognition performance. Instead, recent studies (QualNet-LM and A-SKD) have employed an additional module (decoder or attention) to extract the teacher network's knowledge and transfer it to the student network more concisely. Although they achieved significant improvements in LR face recognition performance on various benchmarks, they could not transfer the entire knowledge obtained from the teacher network's raw features. 

\textbf{Types of Feature Knowledge Distillation.} From this perspective, we analyzed various approaches to feature KD and identified the optimal distance measures for aggregating features of different resolutions. We divided the feature vector ($\mathbf{f}$) into two components: norm ($\lvert\lvert \mathbf{f} \rvert\rvert_p$) and direction ($\frac{\mathbf{f}}{\lvert\lvert \mathbf{f} \rvert\rvert_p}$). Based on this, we defined three types of feature KD:

\begin{itemize}
    \item FitNet~\cite{fitnet}: This approach distills the entire feature using the distance measure $\lvert\lvert \mathbf{f}_T - \mathbf{f}_S \rvert\rvert_p$.

    \item Norm-KD: This approach only distills the norm components using the distance measure $\lvert\lvert (\lvert\lvert \mathbf{f}_T \rvert\rvert_p - \lvert\lvert \mathbf{f}_S \rvert\rvert_p) \rvert\rvert_p$.

    \item F-SKD: This approach only distills the direction components via similarity loss $(1 - \frac{\mathbf{f}_T}{\lvert\lvert \mathbf{f}_T \rvert\rvert_p} \cdot \frac{\mathbf{f}_S}{\lvert\lvert \mathbf{f}_S \rvert\rvert_p})$.
\end{itemize}

In each of the above approaches, $\mathbf{f}_T$ and $\mathbf{f}_S$ represent the features of the teacher and student networks, respectively. $p=2$ following the FitNet.

We evaluated the three feature KD approaches on the 4$\times$ downsampled AgeDB-30 dataset, as shown in Table \ref{tab:feat_type}. Our results demonstrated that F-SKD outperformed the other feature-distillation approaches. This indicates that distilling only the direction components through similarity loss is effective for extracting informative features and guiding the LR network to focus on important regions. In contrast, including norm component distillation between the HR and LR networks led to performance degradation.

\begin{table}[t]
\caption{Evaluation results on the 4$\times$ downsampled AgeDB-30, varying with distillation components (norm and direction).}
\vskip 0.1in
\centering
\label{tab:feat_type}
\resizebox{0.55\textwidth}{!}{%
\begin{tabular}{@{}cccc@{}}
\toprule
\multirow{2}{*}{\textbf{Type}} & \multicolumn{2}{c}{\textbf{Component}} & \multirow{2}{*}{\textbf{ACC (\%)}} \\ \cmidrule(lr){2-3}
 & \textbf{Norm ($\lvert\lvert \mathbf{f} \rvert\rvert_p$)} & \textbf{Direction ($\frac{\mathbf{f}}{\lvert\lvert \mathbf{f} \rvert\rvert_p}$)} &  \\ \midrule
Base & - & - & 85.68 \\
FitNet~\cite{fitnet} & \checkmark & \checkmark & 86.05 \\
Norm-KD & \checkmark & - & 85.18 \\
F-SKD (Ours) & - & \checkmark & \textbf{87.42} \\ \bottomrule
\end{tabular}
}
\end{table}

\begin{figure}
  \centering 
  \includegraphics[height=6.3cm]{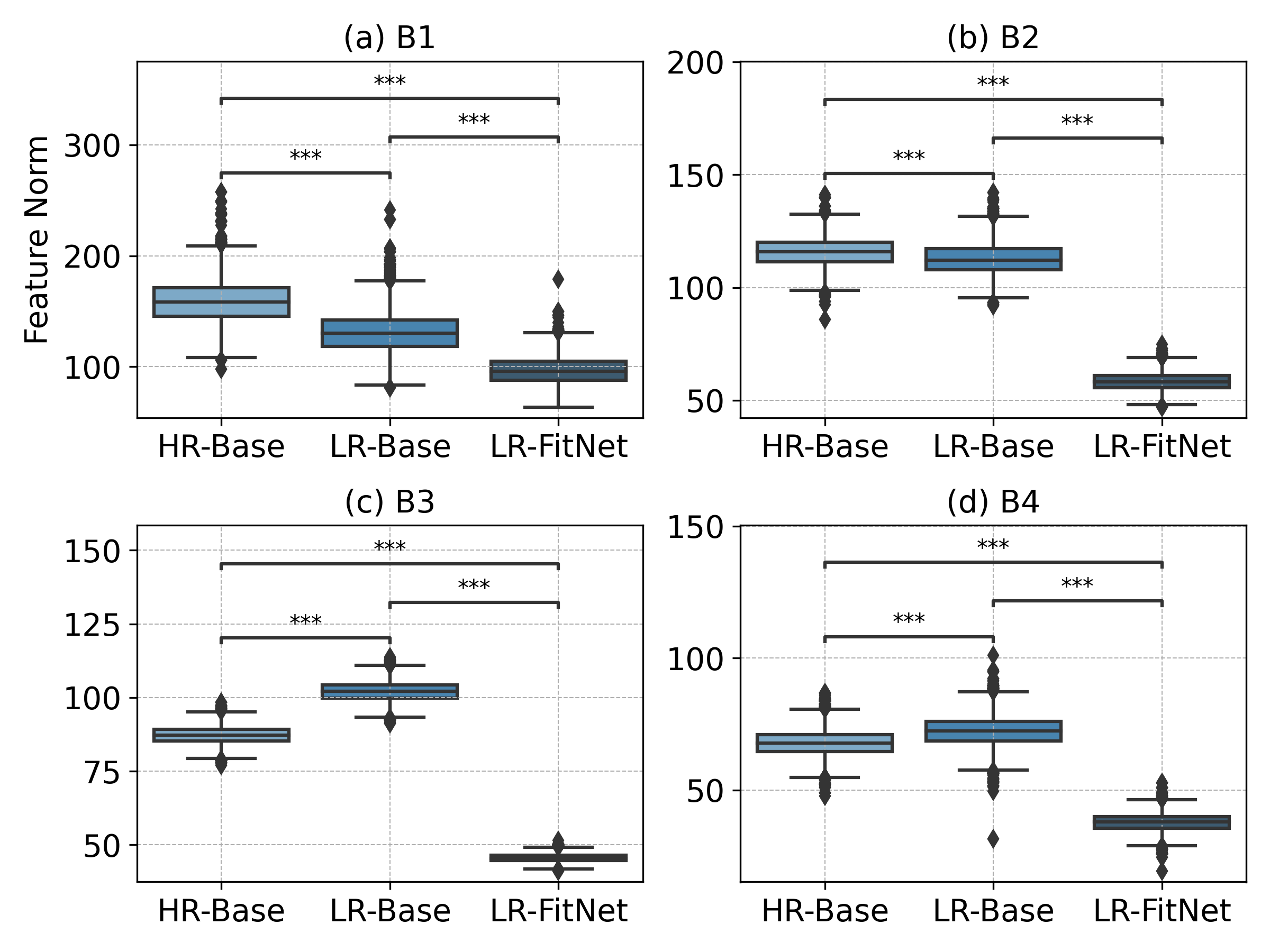}
  \caption{Box plot showing the feature norm for different distillation methods, along with the $t$-test results for the null hypothesis that the features of HR and LR networks are equal. The notation B\{$i$\} represents the $i$th ResNet block. A \textit{p}-value of less than 0.01 (*** in the plot) rejects the null hypothesis, indicating that the feature norm is not statistically equal. \textit{t}-test was performed using the 4$\times$ downsampled AgeDB-30.}
  \label{fig:ttest}
\vskip -0.05in
\end{figure}

\textbf{Statistical Analysis.} In theory, if FitNet~\cite{fitnet} successfully enables the LR network to extract the same features as the HR network, the LR network should achieve the same accuracy as the HR network. However, based on the results of \textit{t}-test, we demonstrated that the average feature norm of the LR network remains statistically different from that of the HR network (\textit{p}-value $< 0.01$), even after applying FitNet, as shown in Figure \ref{fig:ttest}. Additionally, the average feature norm of the LR network decreased across all blocks when FitNet was applied, indicating a potential conflict between the norm component constraints and the task loss associated with face recognition.

\begin{figure}
\centering
\includegraphics[height=7.0cm]{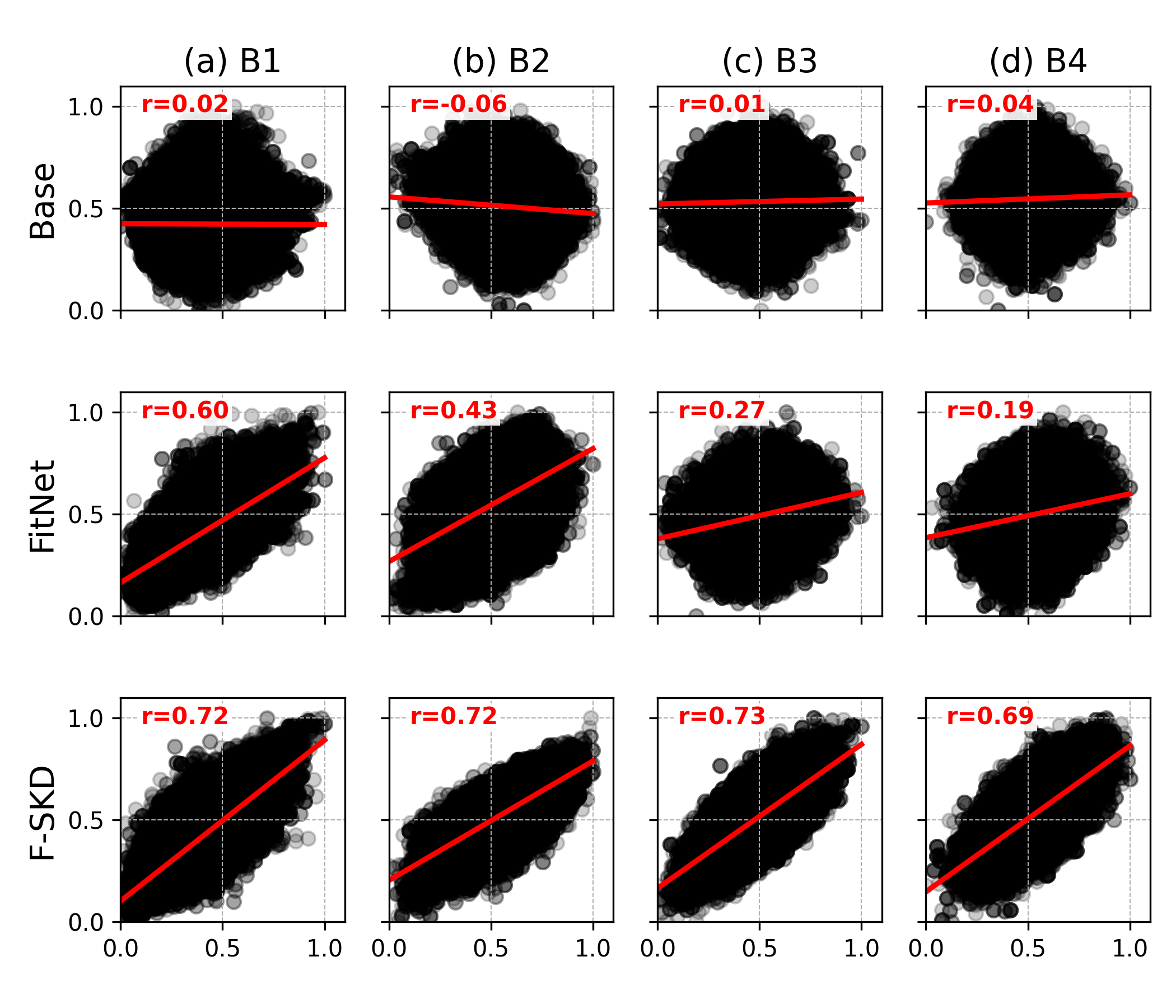}
\caption{Pixel-level Pearson's correlation between HR and LR network features using various distillation methods, where B\{$i$\} represents the $i$th ResNet block. Pearson's correlation coefficient ($r$) measures the strength and direction of the linear relationship between the variables. Pearson's correlation was measured using the evaluation set of AgeDB-30, and the plot shows data points obtained from 1000 randomly selected images.}
\label{fig:correlation}
\vskip -0.05in
\end{figure}

We conducted a Pearson's correlation analysis for the baseline, FitNet, and our proposed F-SKD approaches to investigate the directional component of the features after distillation, as shown in Figure \ref{fig:correlation}. Our analysis revealed that the baseline network did not correlate with the HR network features, indicating that important features for HR recognition were not captured in the LR network. This lack of correlation is a known cause of performance degradation when resolution decreases~\cite{adaface}. In contrast, FitNet yielded a strong positive correlation in the first block's features, but the correlation became weaker in the later blocks. Although there was a slight improvement in LR recognition performance owing to the correlation in the initial blocks, the latter blocks, which extract high-level semantic features, did not resemble the HR network's features. In contrast, our proposed approach achieved a strong positive correlation ($r > 0.6$) between the HR and LR network features for all the blocks. This indicates that our F-SKD approach converged and effectively guides the LR network to focus on informative regions for all the blocks, similar to those of the HR network.

\subsection{Low-resolution Digit Classification}
We extended our approach to the LR digit classification task, which is useful for digit recognition on vehicle registration plates. An evaluation of SVHN~\cite{svhn} shows that our F-SKD method outperforms previous distillation methods with significant margins, as shown in Table \ref{tab:svhn} (+1.73\% compared to RKD, +0.87\% compared to QualNet-LM, and 1.90\% compared to A-SKD). Whereas previous SOTA methods for LR face recognition (QualNet-LM and A-SKD) achieved lower accuracy than conventional KD approaches (FitNet and RKD), our method outperformed all other approaches. This demonstrates that improving the similarity between different resolution features is an effective solution for distilling knowledge from HR to LR networks, not only in face recognition, but also in other general vision tasks.

\begin{table}[t]
\caption{Single-resolution evaluation results of proposed distillation approaches on SVHN. (a) Classification accuracy of FitNet, RKD, QualNet-LM, and F-SKD using iResNet50 as the student network. (b) Classification accuracy of A-SKD and F-SKD using iResNet50+CBAM as the student network.}
\vskip 0.1in
\centering
\label{tab:svhn}
\resizebox{0.6\textwidth}{!}{%
\begin{tabular}{cccc}
\toprule
\textbf{Resolution} & \textbf{Type} & \textbf{Teacher} & \textbf{ACC (\%)} \\ \midrule

\rowcolor{Gray}
(a) &&& \\ \midrule

\multirow{2}{*}{32 $\times$ 32} & \multirow{2}{*}{Base} & iResNet50 & 93.97 \\ \cmidrule(l){3-4} 
    &  & iResNet50+Dec & 93.81 \\ \midrule    
\multirow{6}{*}{8 $\times$ 8} & Base & - & 84.43 \\ \cmidrule(l){2-4} 
& FitNet~\cite{fitnet} & \multirow{3}{*}{iResNet50} & 85.33  \\
& RKD~\cite{rkd} &  &  85.36 \\
& F-SKD (Ours) & & \textbf{86.84} \\ \cmidrule(l){2-4}
& QualNet-LM~\cite{qualnet} & \multirow{2}{*}{iResNet50+Dec} & 85.04 \\
& F-SKD (Ours) & & \textbf{85.78}  \\ \midrule

\rowcolor{Gray}
(b) &&& \\ \midrule

32 $\times$ 32 & Base & iResNet50+CBAM & 93.80 \\ \midrule
\multirow{3}{*}{8 $\times$ 8} & Base & - & 84.25 \\ \cmidrule(l){2-4}
  & A-SKD~\cite{askd} &  \multirow{2}{*}{iResNet50+CBAM} & 84.94 \\
  & F-SKD (Ours) &  & \textbf{86.55} \\ \bottomrule

\end{tabular}
}
\end{table}

\section{Conclusion}

Our study demonstrates that F-SKD is an effective approach for transferring knowledge from an HR network to an LR network. F-SKD distills the directional component of the features by reducing the cosine similarity-based distillation loss, making it the most effective solution for aggregating features of different resolutions. Our experiments show that F-SKD achieves SOTA performance without additional complexity. Statistical analysis tests validated its effectiveness in making LR network features similar to those of the HR network across all blocks. Moreover, F-SKD's efficiency and ease of implementation make it a promising solution for LR recognition applications beyond face recognition in real-world scenarios. In conclusion, our findings demonstrate the potential of F-SKD as a simple yet effective approach to knowledge transfer in LR recognition tasks.

\textbf{Potential Societal Impacts.} The use of deep learning in real-world applications, particularly in face recognition, has significant societal implications. To address the potential privacy concerns associated with facial recognition, we carefully designed our experiments to exclude problematic datasets. Specifically, we excluded MegaFace~\cite{megaface} and MS1MV~\cite{ms1mv}, which have been withdrawn by their authors owing to ethical concerns, as well as IJB-C~\cite{ijbc}, which includes YouTube data in violation of Terms of Service. By prioritizing the minimization of negative societal impacts in our research, we aim to contribute to the development of face recognition technology that can be applied in a responsible and ethical manner. Our study represents a small but important step toward the responsible development and deployment of facial recognition technology, and we hope to inspire further ethical considerations in the field.

\section{Acknowledgements}
This research was financially supported by the Institute of Civil Military Technology Cooperation funded by the Defense Acquisition Program Administration and Ministry of Trade, Industry and Energy of Korean government under grant No. 22-CM-GU-08 as well as by a grant from the Institute of Information and Communications Technology Planning and Evaluation (IITP) funded by the Korean government (MSIT) (No. 2020-0-00857, Development of cloud robot intelligence augmentation, sharing and framework technology to integrate and enhance the intelligence of multiple robots).

\bibliographystyle{unsrtnat}
\bibliography{references}  






\end{document}